\documentclass{article}

\usepackage{microtype}
\usepackage{graphicx}
\usepackage{subfigure}
\usepackage{booktabs} 
\usepackage{multirow}
\usepackage{amsmath}
\usepackage{amssymb}

\usepackage{hyperref}

\usepackage[accepted]{icml2019}

\begin{document}

\twocolumn[
\icmltitle{Deep learning architectures for inference of AC-OPF solutions}




\begin{icmlauthorlist}
\icmlauthor{Thomas Falconer}{ucl}
\icmlauthor{Letif Mones}{invenialabs}

\end{icmlauthorlist}

\icmlaffiliation{ucl}{UCL Energy Institute, London, United Kingdom}
\icmlaffiliation{invenialabs}{Invenia Labs, Cambridge, United Kingdom}

\icmlcorrespondingauthor{Thomas Falconer}{thomas.falconer.19@ucl.ac.uk}

\icmlkeywords{Machine Learning, OPF, Optimization}

\vskip 0.3in
]



\printAffiliationsAndNotice{} 

\numberwithin{equation}{section}
\newcommand{\Constraint}{\mathcal{C}}
\newcommand{\ConstraintEq}{\mathcal{C}^{\textrm{E}}}
\newcommand{\ConstraintIneq}{\mathcal{C}^{\textrm{I}}}
\newcommand{\ConstraintNT}{\mathcal{C}^{\textrm{NT}}}
\newcommand{\Active}{\mathcal{A}}
\newcommand{\Weights}{\Theta}
\newcommand{\Graph}{\mathbb{G}} 
\newcommand{\Node}{\mathcal{V}}
\newcommand{\Real}{\mathbb{R}}
\newcommand{\LoadParams}{\mathcal{X}_{\textrm{load}}}
\newcommand{\GridParams}{\mathcal{X}_{\textrm{all}}}
\newcommand{\Params}{\mathcal{X}}
\newcommand{\Edge}{\mathcal{E}}
\newcommand{\Gen}{\mathcal{G}}

\begin{abstract}
We present a systematic comparison between neural network (NN) architectures for inference of AC-OPF solutions. Using fully connected NNs as a baseline we demonstrate the efficacy of leveraging network topology in the models by constructing abstract representations of electrical grids in the graph domain, for both convolutional and graph NNs. The performance of the NN architectures is compared for regression (predicting optimal generator set-points) and classification (predicting the active set of constraints) settings. Computational gains for obtaining optimal solutions are also presented. 
\end{abstract}

\section{Introduction}
Electricity market dynamics are generally governed by (some form of) Optimal Power Flow (OPF): the computation of (minimal cost) generation dispatch subject to reliability and security constraints of the grid. The classical OPF derivation (AC-OPF) is a non-linear, non-convex constrained optimization problem, which, when integrated with day-ahead unit commitment, forms a Mixed-Integer Program known to be NP-hard. Proliferation of renewable energy resources (e.g. wind and solar) has exacerbated uncertainties in modern power systems, thereby necessitating OPF solutions in near real-time to sustain accurate representation of the system state, preferably combined with probabilistic techniques for modelling uncertainty such as scenario-based Monte-Carlo simulation \citep{Mezghani2019}. This requires sequential OPF solutions at rates unattainable by conventional algorithms.

Typically, AC-OPF is solved using Interior-Point (IP) optimization methods \citep{Wachter2006}, which are computationally expensive given the required evaluation of the Hessian (second-order derivative) of the Lagrangian at each iteration, rendering a time complexity which scales superlinearly with system size. To guarantee suitable rates of convergence, grid operators instead often resort to cheap approximations of OPF, by refactoring the problem using convex relaxations or by utilising the inveterate DC-OPF -- a Linear Program with considerably less control variables and constraints. However, such approximate frameworks are predisposed to sub-optimal estimations of locational marginal prices, engendering wasted emissions-intensive electricity generation -- illustrated by the estimated 500 million metric tons of carbon dioxide emissions per year by virtue of global grid inefficiencies \citep{Surana2019}. Extending the OPF objective to incorporate generator-wise emissions costing furthers the computational burden, hence enabling fast computation of AC-OPF not only facilities renewable energy generation, but is a direct method for climate change mitigation \citep{Gholami2014}.

\begin{figure}[!ht]
    \centerline{\includegraphics[width=\columnwidth]{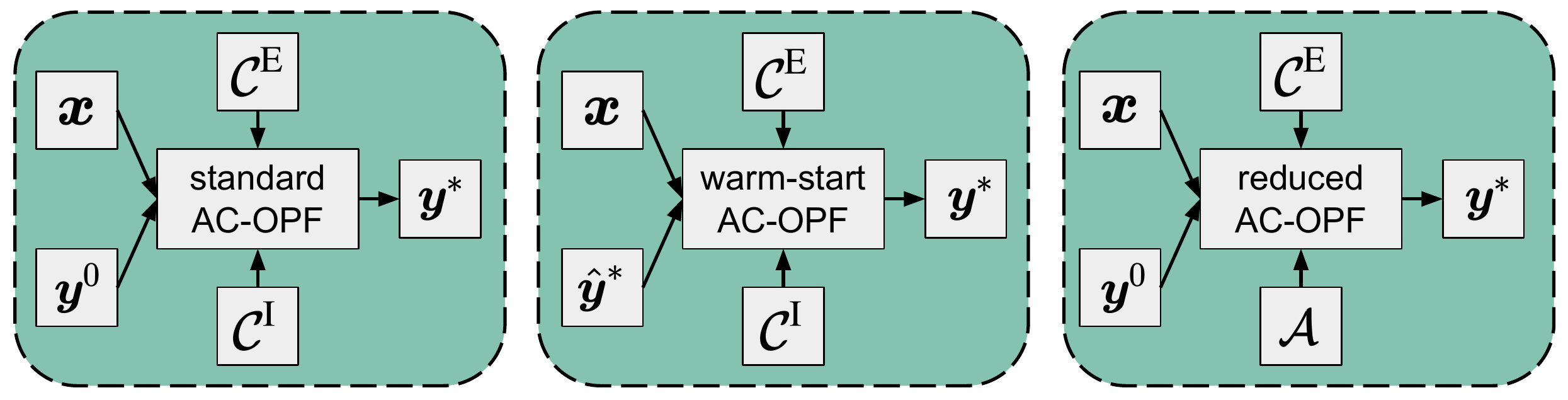}}
    \caption{Strategies for solving AC-OF with IP methods:
    standard (left), warm-start (middle), and reduced (right). $\boldsymbol{x}$ denotes the vector of grid parameters, $\ConstraintEq$ represents the set of equality constraints, $\ConstraintIneq$ and $\Active$ represent the full and active sets of inequality constraints, respectively, and $\boldsymbol{y}^*$ and $\boldsymbol{y}^0$ are the optimal and initial vectors of the primal variables.}
    \label{fig:ml_opf}
\end{figure}

Applications of machine learning (ML) to OPF typically use regression techniques to directly infer the OPF solution from grid parameters, bypassing conventional solvers, thereby shifting the cost from online optimization to offline training \citep{Guha2019}. However, since the optimal point is not necessarily a smooth function of the inputs, this approach requires training on relatively large data sets to obtain acceptable accuracy. Further, there is no guarantee that the inferred optimal solution is feasible (i.e. satisfies all constraints), and violation of important constraints could be catastrophic in the context of power systems. It is more pragmatic therefore to ensure convergence to the optimal solution by instead initialising the IP solver with the inferred optimal solution, a so-called \emph{warm-start} (Figure \ref{fig:ml_opf}, middle panel) \citep{Baker2019}, whereby the regressor is parameterised using a (meta-)loss function corresponding directly to the time complexity to minimise computational burden \citep{Jamei2019}.

Alternatively, the optimal active set \citep{Misra2018} or the binding status of each constraint \citep{Robson2020} can be inferred using classification. Although the number of constraints is exponential in system size \citep{Deka2019}, typically only a fraction constitutes the active set, thus a reduced problem (Figure \ref{fig:ml_opf}, right panel) can be formulated whilst preserving the objective. Security risks associated with false negatives can be avoided by iteratively solving the reduced OPF and adding violated constraints until all those of the full problem are satisfied, hereafter referred to as the \emph{iterative feasibility test}. Since the reduced OPF is much cheaper relative to the full problem, this procedure can in theory be rather efficient -- the computational cost can also be minimised via meta-optimization by directly encoding the time complexity into the (meta-)loss function and optimizing the weights accordingly \citep{Robson2020}.

In each case, it is crucial to use model architectures that maximise predictive performance. The field typically employs fully connected models (FCNNs), however CNNs \citep{Chen2020} and GNNs \citep{Owerko2020} have recently been investigated to exploit spatial dependencies within the electrical grid. This paper addresses contentions throughout literature of the most suitable modelling frameworks for OPF by offering a systematic comparison between these architectures, and between the regression and classification methods. Two input domains are studied: (1) only load variables, to exclusively mimic fluctuations in demand and (2) the entire set of grid parameters, to reflect the seasonality of all grid components and to evaluate the utility of the architectures in (more realistic) high-dimensional scenarios. Experiments are carried out using synthetic grids from Power Grid Lib \citep{pglib2019}. We believe our work can facilitate the transition toward a clean and efficient power grid -- an important contribution to tackling climate change.

\section{Methods}
\begin{figure}[!ht]
    \centerline{\includegraphics[width=0.7\columnwidth]{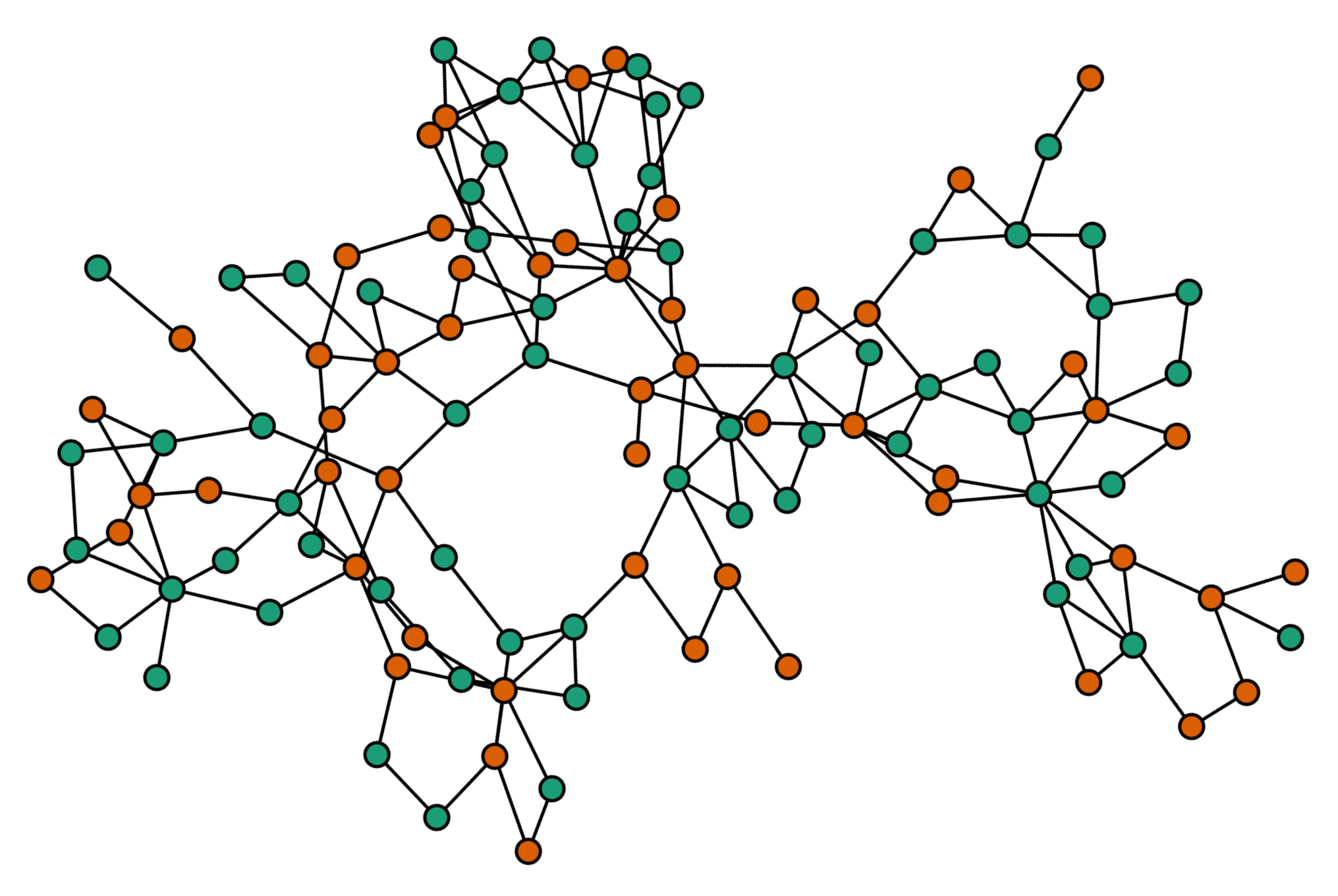}}
    \caption{Graph representation of synthetic grid 118-ieee. Orange and green circles denote generator and load buses, respectively.}
    \label{fig:graph}
\end{figure}

The electrical grid can be represented as an undirected graph $\Graph = (\Node, \Edge)$ with a set of $|\Node|$ number of nodes (buses) and $|\Edge|$ number of edges (transmission lines) such that $\Edge = \{(i, j)\:|\:i\:\textrm{is connected to}\:j;\:i, j \in \Node\}$ (e.g. Figure \ref{fig:graph}).
We define $\Params$ as the set of \emph{control variable types}, which can include the following: active and reactive loads, maximum active and reactive power outputs of generators, line thermal ratings and the reactance and resistance of each transmission line. Hereafter we refer to the two studied domains of the control variable types as: (1) only load variables $\LoadParams$ and (2) all grid parameters $\GridParams$. Further, the above grid variable types can be classified as node $\Params^{\Node}$ or edge $\Params^{\Edge}$ types. The complete set of operational constraints $\Constraint$ is characterised by two distinct subsets: (1) the set of both convex and non-convex equality constraints $\ConstraintEq \subset \Constraint$, which enforce the non-linear AC power flow equations and (2) the set of inequality constraints $\ConstraintIneq \subset \Constraint$, which enforce the operational and physical limits of the system (e.g. lower and upper bounds of generator power injection).

With this notation, AC-OPF can be expressed in the following concise form of mathematical programming:
\begin{equation}
\begin{aligned}
    & \min \limits_{\boldsymbol{y}}\ f(\boldsymbol{x}, \boldsymbol{y}) \\
    & \mathrm{s.\ t.} \ \ c_{i}^{\mathrm{E}}(\boldsymbol{x}, \boldsymbol{y}) = 0 \quad i = 1, \dots, |\ConstraintEq| \\
    & \quad \; \; \; \; \; c_{j}^{\mathrm{I}}(\boldsymbol{x}, \boldsymbol{y}) \ge 0 \quad j = 1, \dots, |\ConstraintIneq| \\
    \label{eq:acopf}
\end{aligned}
\end{equation}
Here, $\boldsymbol{x}$ denotes the vector of grid parameters, $\boldsymbol{y}$ is the vector of voltage magnitudes and active power injections of generators and $f(\boldsymbol{x}, \boldsymbol{y})$ is the objective function -- typically a quadratic or piecewise linear function of the (monotonically increasing) generator cost curves. This formulation (eq. \ref{eq:acopf}) lets us view AC-OPF as an operator that maps the grid parameters to the optimal solution.

For an arbitrary input $\boldsymbol{x}$ the regression objective is to infer the set of optimal primal variables $\boldsymbol{y}^*$, indicative of the minimum information required for the warm-start. Defining $\Gen \subset \Node$ as the set of generators, the regressor learns the \emph{true} AC-OPF mapping defined by the operator $\Omega:\Real^{|\boldsymbol{x}|} \mapsto \Real^{|\Node|+|\Gen|}$ by parameterising a neural network $\hat{\Omega}$ with parameters $\Weights$ and minimising the mean-squared error (MSE) between the ground-truths $\boldsymbol{y}^* = \Omega(\boldsymbol{x})$ and the inferred optimal solution $\boldsymbol{\hat{y}}^* = \hat{\Omega}(\boldsymbol{x}, \Weights)$. 

The classification setting determines the binding status of each constraint in $\ConstraintNT \subset \ConstraintIneq$, the subset of inequality constraints, which change binding status at least once in the training set (hereafter referred to as non-trivial constraints). This facilitates construction of the active set $\Active \subseteq \ConstraintIneq$, the subset of inequality constraints binding at the optimum (such that $\ConstraintEq \cup \Active$ constitute the congestion regime). We define the classifier output as a binary vector representing an enumeration of the set of non-trivial constraints. The classifier therefore learns the mapping $\Psi:\Real^{|\boldsymbol{x}|} \mapsto \{0,1\}^{|\ConstraintNT|}$ provided by the OPF by minimising the binary cross-entropy loss (BCE) between the ground-truths $\Psi(\boldsymbol{x})$ and the predicted probabilities $\hat{\Psi}(\boldsymbol{x}, \Weights)$.

To maintain validity of the constructed data sets, we generated samples within prescribed boundaries around the Power Grid Lib defaults: $\pm15\%$ for nodal active load and $\pm10\%$ for all remaining parameters. On account of demonstrated computational efficiency relative to conventional Uniform scaling methods, we adopt a Random Walk Metropolis-Hastings-based sampling technique whereby steps are taken through normalised space subject to an isotropic Gaussian proposal distribution $\mathcal{N}(\boldsymbol{x}^*, \alpha\textbf{I})$, centred on the current system state $\boldsymbol{x}^*$ with step size $\alpha$, only accepting AC feasible candidates \citep{Falconer2020}. AC-OPF solutions were found using \texttt{PowerModels.jl} \citep{Coffrin2018} in combination with the \texttt{IPOPT} solver \citep{Wachter2006}.

We first construct a baseline FCNN designed to operate on data in vector format, hence lacking sufficient relational inductive bias to exploit any underlying structure. Network topology can be expressed in the graph domain using the (weighted) binary adjacency matrix. For CNNs with only load variables as input, each parameter can be represented by a vector of length $|\Node|$ and subsequently combined into a 3D tensor with dimensions $\Real^{|\Node| \times 1 \times |\Params|}$. For all grid parameters, those relevant to nodes $\Params^{\Node}$ or edges $\Params^{\Edge}$ can be constructed into diagonal or off-diagonal matrices, respectively, such that the input dimensions are $\Real^{|\Node| \times |\Node| \times |\Params|}$.

For GNNs, we investigate both spectral and spatial convolutions, the latter of which are typically constrained to operate on node features and a 1D set of edge features. We overcome this limitation by encoding edge features as node features by concatenating structures akin to that mentioned above, engendering input dimensions of $\Real^{|\Node| \times (|\Params^{\Node}| + |\Params^{\Edge}| |\Node|)}$. Although spatial graph convolutions ordinarily permit multi-dimensional edge features, enhanced performance was observed empirically using an input structure akin to that defined here.

Spectral graph convolutions operate in the Fourier domain; input signals are passed through parameterised functions of the normalised Laplacian, exploiting its positive-semidefinite property. Given this procedure has $\mathcal{O}(|\Node|^3)$ time complexity, we investigate two spectral layers, ChebConv \citep{Kipf2017} and GCNConv \citep{Defferrard2017}, which reduce the computational cost by approximating the kernel functions using Chebyshev polynomials of the eigenvalues up to the $K$-th order, avoiding expensive full eigendecomposition of the Laplacian. GCNConv constrains the layer-wise convolution to first-order neighbours ($K=1$), lessening overfitting to particular localities.

Spatial graph convolutions are instead directly performed in the graph domain, reducing time complexity whilst minimising information loss. For a given node, SplineConv computes a linear combination of its features together with those of its $K$-th order neighbours, weighted by a kernel function using the product of parameterised B-spline basis functions \citep{Fey2018}. The local support property of B-splines reduces the number of parameters, enhancing the computational efficiency of the operator.

Each graph layer (ChebConv, GCNConv and SplineConv) was used to construct a unique GNN, hereafter referred to as CHNN, GCN and SNN, respectively. Parameters were optimized using ADAM \citep{Diederik2014} with learning rate initialised at $\eta=10^{-4}$. Hidden layers were applied with BatchNorm and a ReLU activation function; dropout (with probability of 0.4) was applied to fully-connected layers. CNNs were constructed using \(3\times3\) kernels, \(2\times2\) max-pooling layers, zero-padding and a stride length of 1. For CHNN and SNN, $K$ was set to 5. Hyper-parameters were tuned iteratively, to construct models where for each case, the number of weights was in the same order of magnitude to facilitate a fair comparison of performance. 

\section{Results}
We generated 10k samples for several synthetic grids (Table \ref{tab:samples}). Less unique active sets were uncovered using only load variables since without altering the other grid parameters, changes in system state are much less profound, rendering homogeneity amongst congestion regimes. For the 300-ieee case, this is capped at the number of samples, implying that even 10k samples restricts convergence to the \emph{true} parameter distribution for larger grids. The additional complexity of using all grid parameters is highlighted by the larger input domains $\dim(\boldsymbol{x})$ in addition to the greater cardinalities of non-trivial constraints $|\ConstraintNT|$. Data sets were subsequently split into training, validation and test sets with a ratio of 80:10:10.

\begin{table}[!ht]
\vskip -0.1in
\small
\caption{Grid characteristics and number of unique active sets for different AC-OPF cases, using 10k samples.}
\label{tab:samples}
\vskip 0.1in
\centering
    \resizebox{\columnwidth}{!}{
    \begin{tabular}{lccc|ccc}
    \toprule
    \multirow{2}{*}{Case} & 
    \multicolumn{3}{c|}{Only Load: $\boldsymbol{x} = \boldsymbol{x}_{\textrm{load}}$} &
    \multicolumn{3}{c}{All Parameters: $\boldsymbol{x} = \boldsymbol{x}_{\textrm{all}}$} \\
    \cmidrule(r){2-7}
     & {$\dim(\boldsymbol{x})$} & {$|\ConstraintNT |$} & {\# Active Sets} & {$\dim(\boldsymbol{x})$} & {$|\ConstraintNT|$} & {\# Active Sets}
    \\
     \midrule
     {73-ieee-rts} & 146 & 28 & 1008 & 660 & 60 & 2902 \\
     {118-ieee} & 236 & 31 & 437 & 864 & 66 & 6649 \\
     {162-ieee-dtc} & 322 & 58 & 1805 & 1102 & 92 & 6026 \\
     {300-ieee} & 600 & 143 & 9803 & 1773 & 201 & 10000 \\
     \bottomrule
    \end{tabular}
    }
\vskip -0.1in
\end{table} 

Regression performance is depicted by the average (test set) MSE for five independent runs (Table \ref{tab:reg}); each GNN outperformed FCNN and CNN, in some cases by an order of magnitude. Relative performance is more overt using only load given the lower input dimensionality, however, we still see performance enhancements in the more complex all parameter setting. SNN typically outperforms GCN and CHNN, possibly due to the information loss via spectral node embedding. This, combined with reduced training times in virtue of the computational efficiency of the B-spline kernel functions, alludes to better scaling to larger systems. We observe no significant performance enhancements using CNN, which was expected, since the non-Euclidean data structures are deprived of the geometric priors on which the convolution operator relies. Anomalous cases of reduced error can be attributed to the coincidental unearthing of structural information within the receptive fields when convolving over the (weighted) adjacency matrices.

\begin{table}[!ht]
\vskip -0.1in
\small
\caption{Average test set MSE values of regression models.}
\label{tab:reg}
\vskip 0.1in
\centering
    \resizebox{\columnwidth}{!}{
    \begin{tabular}{llccccc}
    \toprule
    Case ($\boldsymbol{x} = \boldsymbol{x}_{\textrm{load}}$) & & FCNN & CNN & GCN & CHNN & SNN \\
     \midrule
     {73-ieee-rts} & $10^{-4}$ & $6.613$ & $7.625$ & $0.556$ & $0.612$ & $\textbf{0.527}$ \\
     {118-ieee} & $10^{-4}$ & $2.171$ & $3.042$ & $\textbf{0.306}$ & $0.334$ & $0.329$ \\
     {162-ieeet-dtc} & $10^{-3}$ & $9.492$ & $6.026$ & $3.341$ & $3.039$ & $\textbf{2.145}$ \\
     {300-ieee} & $10^{-2}$ & $3.654$ & $5.973$ & $2.283$ & $2.156$ & $\textbf{1.948}$ \\
     \midrule
     Case ($\boldsymbol{x} = \boldsymbol{x}_{\textrm{all}}$)  & & FCNN & CNN & GCN & CHNN & SNN \\
     \midrule
     {73-ieee-rts} & $10^{-3}$ & $4.916$ & $5.241$ & $2.011$ & $1.953$ & $\textbf{1.247}$ \\
     {118-ieee} & $10^{-3}$ & $2.621$ & $3.487$ & $0.396$ & $0.450$ & $\textbf{0.372}$ \\
     {162-ieeet-dtc} & $10^{-2}$ & $2.783$ & $4.585$ & $1.411$ & $1.682$ & $\textbf{1.229}$ \\
     {300-ieee} & $10^{-1}$ & $1.293$ & $1.466$ & $0.723$ & $0.711$ & $\textbf{0.574}$ \\
     \bottomrule
    \end{tabular}
    }
\vskip -0.1in
\end{table} 

Classification performance is reported in terms of average (test set) recall and precision, in addition to BCE, to see how each model balances the trade-off between false positives (FPs) and false negatives (FNs). GNNs again outperform FCNN and CNN in each case, subject to each metric (Table \ref{tab:clf}). Interestingly, we observe a greater precision than recall in virtually every instance, implying the BCE objective is more sensitive to FPs, which is unfavourable given the computational cost of omitting \emph{true} binding constraints from the reduced problem (FNs). This implies the loss function is misaligned with the desired objective and should be altered to bias the reduction of FNs. 

\begin{figure}[!ht]
    \centerline{\includegraphics[width=\columnwidth]{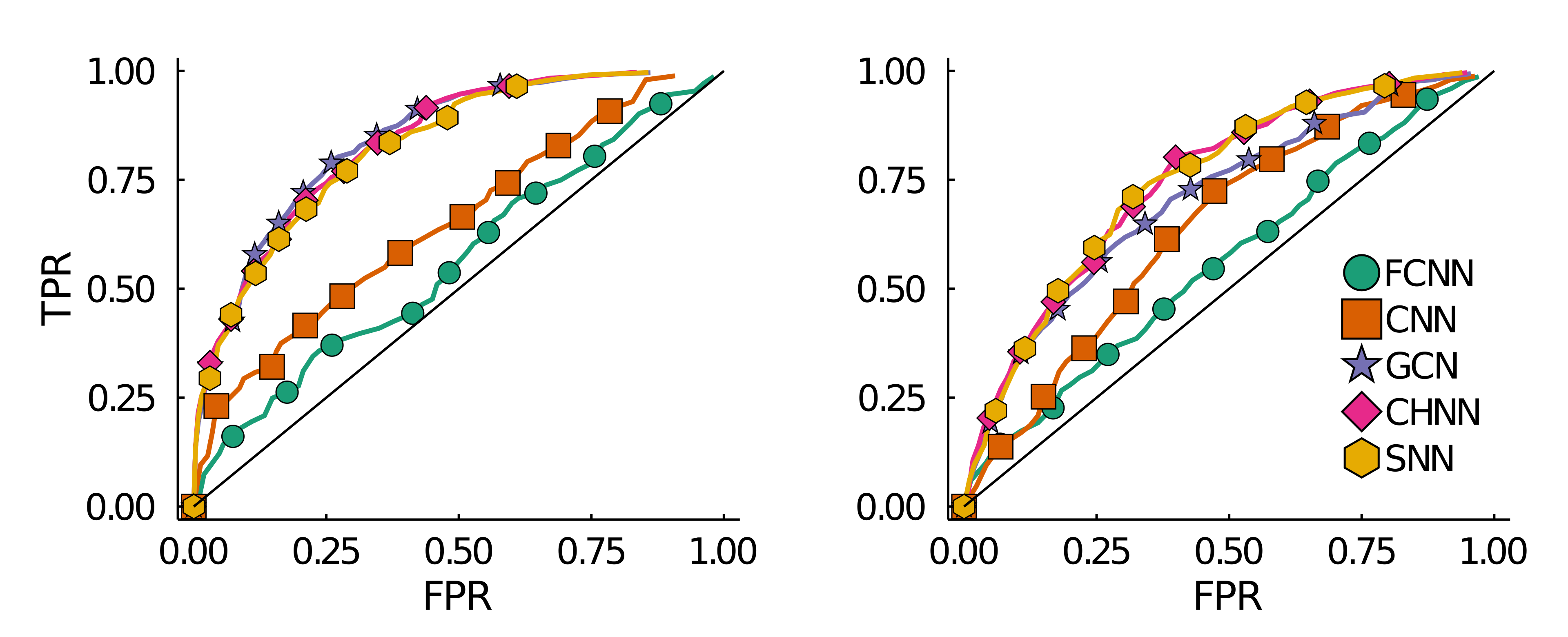}}
    \caption{ROC curves for each classification model (118-ieee system) using only load (left) and all parameters (right) as input.}
    \label{fig:roc}
\end{figure}
Using the 118-ieee system as a case study to visualise the ROC curves for each model (Figure \ref{fig:roc}), we observe that GNNs are superior irrespective of classification threshold. For many of the binary classifications, GNNs demonstrated AUCs surpassing 0.9, whereas FCNN and CNN often reported AUCs less than 0.5, i.e. worse than random. 

\begin{table*}[!ht]
\small
\caption{Average BCE, recall and precision of classification models.}
\label{tab:clf}
\def\na{---}
\centering
    \resizebox{\textwidth}{!}{
    \begin{tabular}{lccccc|ccccc|ccccc}
    \toprule
    \multirow{2}{*}{Case ($\Params = \LoadParams$)} & 
    \multicolumn{5}{c|}{BCE} & 
    \multicolumn{5}{c|}{Recall} & 
    \multicolumn{5}{c}{Precision}\\
    \cmidrule(r){2-16}
     & FCNN  & CNN & GCN & CHNN & SNN & FCNN  & CNN & GCN & CHNN & SNN & FCNN  & CNN & GCN & CHNN & SNN \\
    \midrule
    {73-ieee-rts} & 0.183 & 0.158 & 0.064 & 0.061 & \textbf{0.057} & 0.724 & 0.774  & 0.854 & 0.843 & \textbf{0.865} & 0.928 & 0.935 & 0.964 & 0.946 & \textbf{0.976} \\
    {118-ieee} & 0.193 & 0.179 & 0.041 & \textbf{0.039} & 0.052 & 0.631 & 0.658 & 0.772 & 0.738  & \textbf{0.792} & 0.942 & 0.875 & \textbf{0.957} & 0.929 & 0.946 \\
    {162-ieeet-dtc} & 0.235 & 0.214 & 0.093 & 0.089 & \textbf{0.084} & 0.682 & 0.675 & 0.764 & \textbf{0.767} & 0.761 & 0.773  & 0.782 & 0.857 & \textbf{0.871} & 0.864 \\
    {300-ieee} & 0.157 & 0.163 & 0.111 & \textbf{0.107} & 0.115 & 0.611 & 0.582 & \textbf{0.683} & 0.678 & 0.664 & 0.775 & 0.791 & 0.874 & \textbf{0.882} & 0.879 \\
    \midrule
   \multirow{2}{*}{Case ($\Params = \GridParams$)} & 
    \multicolumn{5}{c|}{BCE} & 
    \multicolumn{5}{c|}{Recall} & 
    \multicolumn{5}{c}{Precision}\\
    \cmidrule(r){2-16}
     & FCNN  & CNN & GCN & CHNN & SNN & FCNN  & CNN & GCN & CHNN & SNN & FCNN  & CNN & GCN & CHNN & SNN \\
    \midrule
    {73-ieee-rts} & 0.263 & 0.247 & 0.115 & 0.123 & \textbf{0.111} & 0.602 & 0.597 & 0.642 & 0.699 & \textbf{0.724} & 0.893 & 0.874 & 0.89  & 0.892 & \textbf{0.921} \\
    {118-ieee} & 0.224 & 0.206 & 0.104 & 0.101 & \textbf{0.098} & 0.532  & 0.614 & \textbf{0.685} & 0.674 & 0.676 & 0.846 & 0.833 & 0.873 & \textbf{0.891}  & 0.86 \\
    {162-ieeet-dtc} & 0.187 & 0.194 & 0.136 & \textbf{0.127} & 0.131 & 0.491 & 0.482 & 0.538 & 0.523 & \textbf{0.552} & 0.823 & 0.835 & 0.879 & \textbf{0.886} & 0.873 \\
    {300-ieee} & 0.195 & 0.201 & 0.128 & 0.121 & \textbf{0.119} & 0.587 & 0.583 &  0.635 &  0.638 & \textbf{0.642} & 0.794 & 0.801 & 0.847 & 0.845 & \textbf{0.859} \\
    \bottomrule
    \end{tabular}
    }
\end{table*}

Finally, we report the average computational gain of the test set realised for the warm-start and reduced problem, using all grid parameters as a case study (Table \ref{tab:gain}). In both settings, the models that provide more accurate inference (GNNs) engender superior results. Moderate gains were expected since differentiating the non-convex equality constraints remains to greatest computational expense in both the warm-start approach and reduced formulation \citep{Robson2020}. Moreover, in the regression setting, only the primal variables are initialised, therefore a minimum number of iterations is still required for the duals to converge i.e. there is an empirical upper bound.

Given the complexity of the 162-ieee-dtc system, it is probable that the primals were near the feasible boundary, rendering an ill-conditioned system thus constraining the solver to small step sizes, hence the negative gain. The similar or greater performance of the reduced problem relative to the warm-start implies that this method is preferable overall, as we could achieve greater gains merely by reducing the number of FNs (thereby reducing the cost of the iterative feasibility test) with a more sophisticated objective function.

\begin{table}[!ht]
\small
\caption{Average computational gain for direct (warm-start) and indirect (reduced problem) approaches.}
\label{tab:gain}
\centering
    \resizebox{\columnwidth}{!}{
    \begin{tabular}{lrrrrr}
    \toprule
    Case (direct)  & \multicolumn{1}{r}{FCNN} & \multicolumn{1}{r}{CNN} & \multicolumn{1}{r}{GCN} & \multicolumn{1}{r}{CHNN} & \multicolumn{1}{r}{SNN} \\
    \midrule
    {73-ieee-rts} & $17.58$ & $15.76$ & $\textbf{21.65}$ & $21.18$ & $21.43$ \\
    {118-ieee} & $14.38$ & $15.92$ & $\textbf{17.27}$ & $16.57$ & $17.09$ \\
    {162-ieee-dtc} & $-133.74$ & $-125.79$ & $-96.76$ & $-93.79$ & $\textbf{--90.64}$ \\
    {300-ieee} & $12.20$ & $11.67$ & $17.31$ & $\textbf{17.56}$ & $16.38$ \\
    \midrule
     Case (indirect)  & \multicolumn{1}{r}{FCNN} & \multicolumn{1}{r}{CNN} & \multicolumn{1}{r}{GCN} & \multicolumn{1}{r}{CHNN} & \multicolumn{1}{r}{SNN} \\
    \midrule
    {73-ieee-rts} & $-13.82$ & $-11.91$ & $-4.26$ & $-2.60$ & $\textbf{--1.31}$ \\
    {118-ieee} & $21.17$ & $22.91$ & $\textbf{27.06}$ & $26.44$ & $25.75$ \\
    {162-ieee-dtc} & $-38.29$ & $-25.08$ & $-10.87$ & $\textbf{--9.10}$ & $-12.26$ \\
    {300-ieee} & $13.24$ & $12.83$ & $\textbf{17.84}$ & $16.90$ & $17.73$ \\
    \bottomrule
    \end{tabular}
    }
\end{table}

\section{Conclusion}
Our systematic comparison of NN architectures for direct and indirect inference of AC-OPF solutions, as well as methods for augmenting the IP solver, has highlighted the importance of predictive accuracy in reducing the computational cost of AC-OPF with ML, essential for minimising emissions in the electricity market. We demonstrated the utility of explicitly incorporating network topology in the learning process using GNNs and concluded the gains of the reduced problem allude to better scaling to larger grids. In future work, we will investigate more sophisticated objectives to bias the optimization towards a reduction in FNs, and leverage the capacity of GNNs to supplement the meta-optimization technique proposed by \citet{Robson2020}.

\bibliography{abstract}
\bibliographystyle{icml2019}

\end{document}